# Reverse Refinement Network for Narrow Rural Road Detection in High-Resolution Satellite Imagery

Ningjing Wang, Xinyu Wang*, *Member, IEEE*, Yang Pan, Wanqiang Yao and Yanfei Zhong, *Senior Member, IEEE*

*Abstract*—The automated extraction of rural roads is pivotal for rural development and transportation planning, serving as a cornerstone for socio-economic progress. Current research primarily focuses on road extraction in urban areas. However, rural roads present unique challenges due to their narrow and irregular nature, posing significant difficulties for road extraction. In this article, a reverse refinement network (R2-Net) is proposed to extract narrow rural roads, enhancing their connectivity and distinctiveness from the background. Specifically, to preserve the fine details of roads within high-resolution feature maps, R2-Net utilizes an axis context aware module (ACAM) to capture the long-distance spatial context information in various layers. Subsequently, the multi-level features are aggregated through a global aggregation module (GAM). Moreover, in the decoder stage, R2-Net employs a reverse-aware module (RAM) to direct the attention of the network to the complex background, thus amplifying its separability. In experiments, we compare R2-Net with several state-of-the-art methods using the DeepGlobe road extraction dataset and the WHU-RuR+ global large-scale rural road dataset. R2-Net achieved superior performance and especially excelled in accurately detecting narrow roads. Furthermore, we explored the applicability of R2-Net for large-scale rural road mapping. The results show that the proposed R2-Net has significant performance advantages for large-scale rural road mapping applications.

## I. INTRODUCTION

Rural roads, as important infrastructure, are not only crucial for residents' travel but also play a key role in agricultural development and the enhancement of the rural economy[1]. However, existing road mapping is often inadequate, particularly in many unpaved rural and agricultural areas[2]. For instance, current OpenStreetMap (OSM) data primarily focuses on urban areas, resulting in a serious lack of coverage for rural road networks, which hinders effective infrastructure planning, traffic management, and environmental monitoring in these rural regions[3]. Therefore, automating the extraction of rural roads is of great significance for promoting agricultural development and improving the rural economy.

In recent years, extracting roads from remote sensing imagery has emerged as a significant research topic [4]. Methods in this field can be primarily categorized into traditional approaches and deep learning approaches. Traditional methods rely on expert knowledge to design distinguishing features for road extraction, which typically include textural features [5-7], geometric features [8-10], and spectral features [11-13]. However, traditional methods have significant drawbacks, primarily characterized by low automation and poor generalization capabilities. With the rapid development of convolutional neural networks [14] in the field of image processing, road extraction methods based on deep learning have gradually become the mainstream method. The initial applications of deep learning focused on convolutional neural networks for segmentation but faced challenges related to high computational costs and limited field of view [15-17]. To address these issues, researchers have utilized encoder-decoder structures to effectively integrate multi-scale information and enhance feature representation [18-21]. Among them, some methods have adopted refinement strategies to merge contextual information, while others have focused on improving boundary detection and maintaining continuity in road extraction. Some other methods use a two-stage post-processing approach to optimize the segmentation results [22-26]. These methods are designed based on the statistical laws of road structures to enhance the results in a heuristic way. In addition, recent studies leveraging generative adversarial networks have further improved segmentation quality through adversarial learning techniques [27, 28]. Overall, these deep learning methods have played a significant role in advancing road extraction technologies.

Although these deep learning-based methods have achieved relatively ideal results in the field of road extraction, they still face the following challenges in the practical application of rural road extraction: 1) The roads are narrow. Due to the typically narrow nature of rural roads and significant width variations, the features of narrow roads appear particularly weak against complex backgrounds. Existing methods often extract only limited spatial context information in the last layer of the encoder. As the network depth increases, the spatial resolution of the feature map gradually decreases, causing the network to lose specific detailed information in high-level

This work was supported by the National Key Research and Development Program of China under Grant No. 2022YFB3903502, in part by the National Natural Science Foundation of China under Grant No. 42325105. (*corresponding author: Xinyu Wang, E-mail: wangxinyu@whu.edu.cn)

Ningjing Wang and Wanqiang Yao are with the School of Surveying and Mapping, Xi'an University of Science and Technology, Xi'an 710054, China (e-mail: wangningjing@stu.xust.edu.cn; sxywq@xust.edu.cn).

Xinyu Wang are with the School of Remote Sensing and Information Engineering, Wuhan University, Wuhan 430079, China (e-mail: wangxinyu@whu.edu.cn).

Yang Pan and Yanfei Zhong are with the State Key Laboratory of Information Engineering in Surveying, Mapping and Remote Sensing, Wuhan University, Wuhan 430079, China (e-mail: panyang@whu.edu.cn; zhongyanfei@whu.edu.cn).

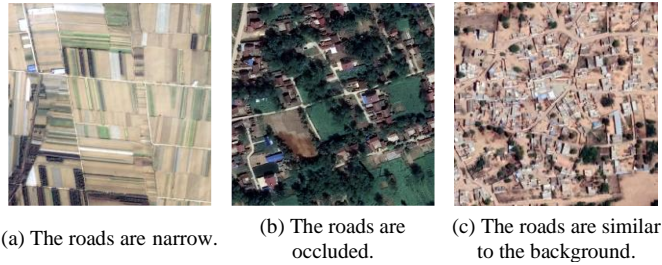

(a) The roads are narrow.  (b) The roads are occluded.  (c) The roads are similar to the background.

Fig. 1. The challenges of rural road extraction.

abstraction, especially when dealing with narrow rural roads. This results in existing methods missing a large number of rural roads during extraction, thereby affecting the integrity of the road network. 2) The roads are occluded. The large number of occluders, such as trees and buildings, in rural areas causes discontinuities in road extraction results. Although existing methods use techniques such as dilated convolution to enhance the receptive field to capture contextual information, or use post-processing methods to fill gaps in the extraction results, these measures are often less effective for narrow rural roads. Especially in rural environments, the diversity and irregularity of occlusions make it difficult for the model to identify continuous roads effectively. 3) The roads are similar to the background. Due to the presence of numerous unpaved roads in rural areas, particularly in the context of natural materials such as soil, grass, or gravel, the texture of the roads often resembles that of the surrounding environment, leading to blurred boundaries in road extraction. Existing methods often lack sufficient feature expression capabilities when dealing with these subtle differences, particularly when lighting conditions change or the background is complex. Existing models may find it difficult to accurately capture the unique characteristics of the road, leading to blurred or jagged road boundaries and shapes, further affecting the accuracy and completeness of the extraction results.

In this article, to address the above issues, we propose a reverse refinement network, namely R2-Net, which consists of three main stages: 1) a feature extraction stage; 2) a positioning stage; and 3) a refinement stage, and works in a coarse-to-fine manner. Specifically, the first two stages are responsible for locating and distinguishing rough position information for roads with weak features, and the third stage utilizes the cooperation mechanism between foreground and background to gradually refine the results. This study extends from our previous work[29]. The proposed R2-Net method obtained a performance increment on two public datasets (WHU-RuR+ and DeepGlobe), confirming the superiority of R2-Net over the other state-of-the-art road extraction methods. In addition, to evaluate the effectiveness of the developed method in rural road extraction, we selected Hubei province in China as an auxiliary region to evaluate the model's generalizability and transferability through large-scale validation. The main contributions of this paper are summarized as follows.

1) A novel method—R2-Net—is proposed for extracting narrow rural roads from high-resolution satellite imagery, focusing on capturing and distinguishing these roads in complex rural backgrounds by effectively integrating global contextual features and enhancing the representation of both foreground and background.

2) Two modules—ACAM and RAM—are proposed, among which ACAM uses multi-branch and dilated convolutional layers to capture detailed multi-scale contextual information of rural roads, while RAM enhances the representation of rural roads by refining road background features, thereby improving its separability from complex backgrounds.

3) A series of experiments and analyses are conducted to evaluate the performance of the proposed R2-Net in the rural road extraction task. At the same time, a large-scale validation was carried out in rural areas, and the results showed that the proposed method has significant performance advantages in rural road mapping.

The remainder of this paper is organized as follows. Section II introduces the related works. Section III gives the details of the proposed R2-Net. Section IV presents the experimental results and analysis. Finally, Section V concludes the article.

## II. RELATED WORKS

Over the past decade, to meet the requirements of road extraction in large-scale and complex backgrounds, various deep learning-based methods have been proposed for road extraction, due to their strong feature expression ability. Among these, segmentation-based road extraction methods can be mainly categorized into three prominent paradigms: patch-based CNNs, encoder-decoder architectures, and post-processing methods. In this section, we will introduce these three categories of deep learning methods for road extraction in detail.

### A. Methods Based on Patch-CNNs

The initial application of deep learning to road extraction was based on an image block convolutional neural network (CNN) [16] for segmentation. The process of extracting roads from remote sensing images using a patch-based CNN involves several key steps. Initially, the image and segments are divided into patches. These patches are then fed into the CNN to extract features and identify those containing road information. Finally, the identified road patches are aggregated to produce the complete road network. Since the fully connected layer can only accept a fixed-size image input, the architecture cuts the image into fixed-size blocks and inputs them into the network model for learning. The encoder part can be replaced by any backbone deep convolutional network, such as VGG[30], ResNet[31], etc. For example, in [15], the authors proposed a deep network based on restricted Boltzmann machines (RBMs) to segment road areas from high-resolution remote sensing images. In [17], the authors combined low-level features (i.e., asymmetry and compactness) of neighboring roads and buildings with CNN features in a post-processing stage. In contrast, in [32], a coarse-to-fine road extraction strategy combining grayscale distribution and structural features was proposed. In terms of network training, the channel-wise inhibited softmax (CIS) function was proposed in [33] to improve training effectiveness. Additionally, the impact of different patch sizes and input image resolutions on segmentation accuracy was explored in [34], which introduced the multi-scale collective fusion (MSCF) method to extract information from multiple resolutions. Unlike the aforementioned surface extraction methods, a patch-based CNN model was utilized to extract road centerlines from high-resolution remote sensing images [35], and line integral convolution (LIC) was designed to optimize the extracted road network. For road centerline extraction as well, a patch-based four-stage method was proposed in [36], which employed Gabor filtering models and multi-directional non-maximum suppression techniques for extraction. Although the above method has achieved good results in road extraction, since these methods need to cut the image into many patches for inference and then splice the prediction results, there is always the problem of boundary discontinuity. In addition, the repeated operation of image patches and the existence of fully connected layers will lead to low computational efficiency.

*B. Methods Based on Encoder–Decoder*

The encoder-decoder architecture is one of the mainstream technologies in current deep-learning road extraction methods. It is highly favored for its excellent context capture ability and detail preservation performance. This method extracts features from the input image through an encoder and then uses a decoder to gradually restore the spatial resolution to achieve pixel-level classification. In this process, the convolutional layer of the encoder is responsible for extracting multi-level features, while the decoder restores the feature map to the size of the original image through upsampling operations. In recent years, researchers have proposed a variety of improved encoder-decoder models for road extraction tasks and adopted a variety of effective strategies. Among them, some methods use multi-scale context information fusion strategies to enhance the efficiency of road feature extraction. For example, the D-LinkNet [19]uses the LinkNet [37] as the baseline model and uses dilated convolutions [38, 39] in the middle part to expand the model receptive field, winning the DeepGloble road extraction challenge. To enable the encoder-decoder structure to effectively fuse global context features, GAMSNet [20] uses global awareness operations to capture spatial context dependencies and inter-channel dependencies. Similarly, GCB-Net [40] also introduces global context-aware blocks in the network to capture the global information of the road. To achieve feature fusion, HCN [41] consists of three sub-networks and allows the extraction of road features at different granularities. Although these methods have achieved good results, they focus on segmenting the entire road area and ignore the constraints of the region boundary. To this end, BT-RoadNet [42] uses a boundary- and topology-aware road extraction network to solve the problem of inaccurate road boundaries under extreme conditions. Although these methods improve the extraction effect to a certain extent, they do not fully capture the relationship between regions and boundaries. Based on this, some studies have designed network architectures specifically for road extraction tasks. For example, DDU-Net [43] introduced a dual decoder U-Net, which promotes multi-scale feature fusion by combining a new dilated convolution attention module. In AD-RoadNet [44], a hybrid receptive field module and a topological feature representation module are used to more effectively capture road details. To further improve the road extraction performance, some methods are optimized by customizing the loss function. For example, in [45], the authors proposed a topology-aware loss specifically for linear features, using a novel iterative refinement method to identify high-order topological features. At the same time, the structural similarity (SSIM) loss function [46] was proposed to improve the clarity of the extraction results, while in [47], the direction mapping was implemented through the angle operator loss function to enhance the road direction recognition ability. To enhance the topological continuity, DSCNet [48] proposes a continuity constraint loss function based on persistent homology to better constrain the topological continuity of the segmentation. Although these methods have made some progress in improving road extraction, they still have problems in handling complex scenes and coping with noise interference.

*C. Methods Based on Post-processing*

The post-processing optimization method improves the accuracy of road extraction and enhances the integrity and continuity of the road network by optimizing the initial extraction results. These methods usually combine image processing techniques, such as morphological operations or conditional random fields, to eliminate discontinuous parts of the road based on the CNN network to extract roads. For example, RDRCNN [23] adopts a two-stage road extraction method, which first uses a deep residual convolutional neural network for initial road extraction, and then uses mathematical morphology and tensor voting algorithms to improve performance. To improve the accuracy and connectivity of road extraction, in [24], the authors proposed an inner convolutional integrated encoder-decoder network and post-processed the directional conditional random field. Different from the above methods, RoadCorrector [25] uses a structure-aware post-processing method to significantly improve the extraction accuracy and topology accuracy by connectivity refinement and topology correction modules. Although these methods have improved the road connectivity problem to a certain extent, their effectiveness is often limited by the quality of the initial road segmentation results, and for structures such as narrow rural roads or tortuous paths, post-processing is often difficult to effectively restore and optimize.

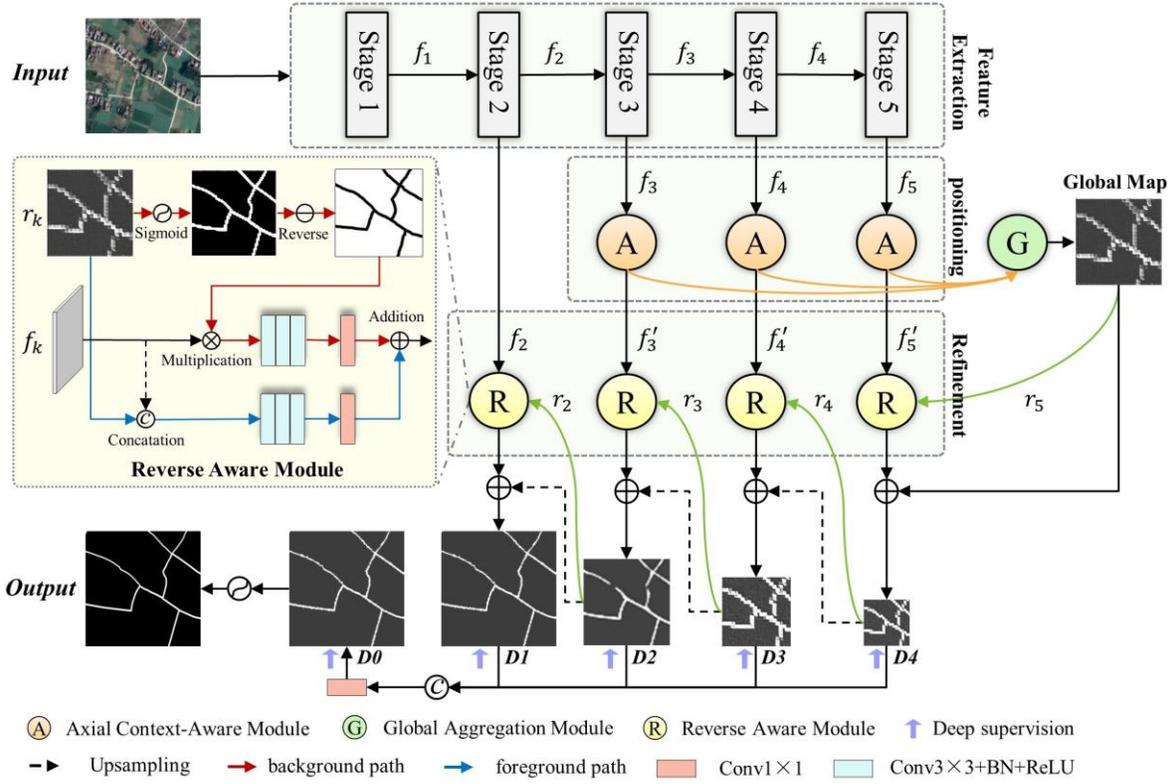

Fig. 2. The R2-Net architecture, where A, G, and R refer to the axis context aware module, the global aggregation module, and the reverse-aware module, respectively. f1–f5 represent the features of different resolutions output from the encoder. D0–D4 represent the in-depth supervision from the different side output layers, respectively. The network consists of three main stages: feature extraction stage, positioning stage, and refinement stage. Specifically, the feature extraction stage is mainly responsible for feature extraction, the positioning stage is responsible for locating and distinguishing the rough position information of the roads, and the refinement stage uses the cooperation mechanism between the foreground and the background to gradually refine the results.

## III. THE REVERSE REFINEMENT NETWORK

In this section, the details of the proposed R2-Net are introduced. To begin with, an introduction is provided to the general framework of R2-Net. Subsequently, the axis context aware module used for extract positional context features from multiple levels of the high-level space. In addition, the global aggregation module is introduced to generate a global feature map of the road network. Then, we introduce the reverse-aware module for enhancing the distinction between foreground road and background. Finally, the loss function used for network training is described in detail.

### A. Overall Architecture

Reverse Refinement Network (R2-Net) is proposed to solve the problem of inaccurate narrow road extraction in rural road extraction. Our motivation stems from the fact that, during the road annotation process, annotation workers first roughly locate the road and then accurately outline its contour mask based on local features. Therefore, we argue that delimiting areas and defining boundaries are the two key features that can be used to distinguish non-roads from roads. Based on this, R2-Net consists of three main stages: 1) **Feature extraction**; 2) **Positioning**; and 3) **Refinement**. R2-Net uses ResNeSt-50[49] as the encoder for the feature extraction and first uses an axis context aware module (ACAM) to extract positional context features from multiple levels of the high-level space. These high-level context features are aggregated using a global aggregation module (GAM) to generate a global feature map of the road network. However, this process only obtains rough position information for the roads. To further refine the structural and textural details of the roads, we use this feature map as the initial guidance area and input it into the reverse-aware module (RAM), which promotes the feature representation of the foreground and background and refines some of the inconsistent predictions by repeatedly using the cooperation mechanism between foreground and background. This process gradually improves the ability to distinguish weak road features. This allows the network to capture and identify narrow roads more accurately. At the same time, the entire network adopts a deep supervision strategy, and multiple side output layers supervise the backbone network at the same time, so that the shallow layers can be more fully trained. The proposed architecture is shown in Fig. 2. The implementation details of the ACAM, GAM, and RAM are described in detail below.

### B. Axis Context Aware Module

Research on rural road extraction is faced with two main challenges. Firstly, the current mainstream road extraction networks usually adopt U-Net or U-Net-like structures (e.g., ResUNet [19], etc.), and the models follow the encoder-decoder framework. However, these models only extract limited spatial context information at the last layer of the encoder. As the depth of the network increases, the spatial resolution of the feature map gradually decreases, causing the network to lose some specific detailed information in high-level abstraction, especially when processing narrow rural roads. Secondly, for long-span features such as roads, the traditional convolution operations are limited by the convolution kernel size and stride, and cannot effectively capture the correlation between long-distance pixels, especially for large-scale images.

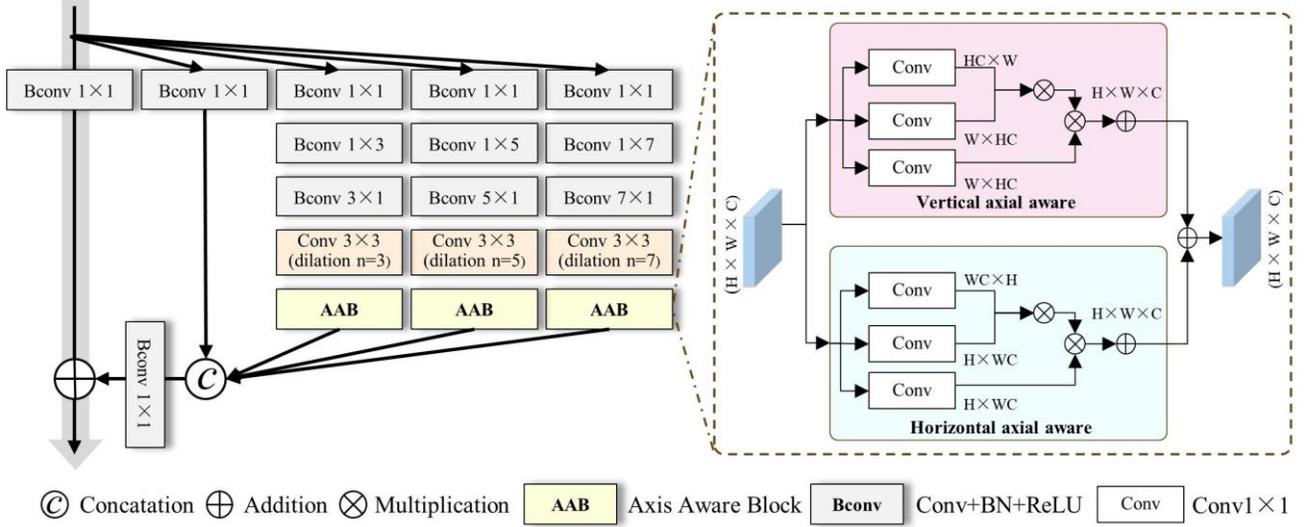

Fig. 3. The axis context aware module (ACAM). The inputs are the features of different resolutions output by the encoder.

Neuroscience experiments have confirmed that the population receptive fields (pRFs) of varying sizes in the human visual system contribute to emphasizing the region near the retinal fovea, which is sensitive to slight spatial shifts [50]. On the other hand, the self-attention mechanism has been proven to be able to achieve the association between the target pixel and any pixel. However, the current self-attention methods (e.g., non-local attention [51]) are computationally intensive, especially when processing large-size feature maps, and their computational efficiency is very low. Therefore, we urgently need a method that can integrate global information without losing detailed information and reduce the amount of calculation to improve the efficiency and accuracy of rural road extraction.

To better capture the detailed information and extensive spatial relationships of roads, we propose the ACAM, as shown in Fig. 3. Specifically, the ACAM includes multi-branch convolutional layers with different core numbers and dilated convolutional layers with different expansion rates. With this design, the module can capture more contextual information over a wider range while maintaining the same number of parameters. At the same time, to refine the feature representation more effectively at the global scale, we introduce an axis-aware block (AAB) at each scale, which performs non-local operations along the horizontal and vertical axes, allowing the network to better understand the local features and enhance global perception. By using the ACAM at multiple levels throughout the network, the network can better adapt to road features of different scales and shapes, thereby enhancing the model's ability to understand complex road structures. At the same time, the introduction of the ACAM does not significantly increase the number of parameters of the model, thus effectively improving the computational efficiency and inference speed of the model.

The ACAM is made up of five branches, denoted as $\{b_k, k = 1, \ldots, 5\}$. Within each branch, the initial convolutional layer has a $1 \times 1$ kernel to downscale the channel size to 64. Following this, there are two additional layers: a convolutional layer and a convolutional layer with a specific dilation rate for when $k > 1$ [39]. Subsequently, each branch sequentially applies an AAB. The attention weights for the vertical and horizontal axes are computed in parallel to capture the relationships between the different positions in the input feature map. The output feature maps from each branch are then further reduced to 64 channels using a convolutional layer and are residual-connected with the original input feature map. Finally, a rectified linear unit (ReLU) activation function is applied to obtain the final encoder output feature. Furthermore, some studies (e.g., Inception-V3 [52]) have shown that a standard convolution operation of size $(2i-1) \times (2i-1)$ can be decomposed into two steps, using a $(2i-1) \times 1$ convolution kernel and a $1 \times (2i-1)$ convolution kernel. This decomposition can improve the reasoning efficiency without reducing the representation capability. In summary, the ACAM enhances the receptive field by adding an extra branch with a larger expansion rate, compared to the standard receptive field block structure. It replaces the standard convolution with two asymmetric convolutional layers and incorporates an AAB to effectively extract both the global relationships and local features.

## C. Global Aggregation Module

Low-level features in the shallow layers preserve the spatial details for constructing object boundaries, while high-level features in the deep layers retain semantic information for locating objects [53]. In the rural road extraction task, first locating the approximate location of the roads is a critical step, especially when there are many narrow roads. Therefore, we classify the extracted features into low-level $\{f_1, f_2\}$ and high-level $\{f_3, f_4, f_5\}$ features, and consider aggregating the high-level features to obtain the global feature map needed to locate roads. However, when aggregating multiple features, two key issues remain: 1) how to maintain semantic consistency within a single layer; and 2) how to connect the context across layers. To solve these problems, we propose the GAM, which is inspired by the cascaded partial decoder (CPD) [54].

The GAM combines high-level features at three different scales from the positioning stage through a series of dense connection strategies to obtain the global feature map. Specifically, high-level features $f_3'$ and $f_4'$ are obtained from the positioning stage. Element-level multiplication operations are then performed on the feature maps to enhance the correlation between features and reduce the gap between multi-layer features. For the highest-level feature $f_5'$, both a 1× upsampling and convolution operation and a 2× upsampling and convolution operation are performed. Similarly, for feature $f_4'$, a 1× upsampling and convolution operation is performed, followed by element-wise multiplication with feature $f_3'$. Finally, an upsampling concatenation strategy is employed to integrate the features across multiple hierarchical levels. This operation is defined as follows:

$$f_4'' = Bconvs(f_4' \odot Up_\uparrow^2(f_5')) \quad (1)$$

$$f_3'' = Bconvs(f_3' \odot Up_\uparrow^4(f_5') \odot Up_\uparrow^2(f_4')) \quad (2)$$

$$f_g = Conv(Bconvs(Cat(f_3'', Up_\uparrow^2(Cat(f_4'', Up_\uparrow^2(f_5')))))) \quad (3)$$

where $f_4''$ represents the output of the first dense connection, which is the result of the upsampling and convolution of the input $f_4'$ and $f_5'$. Similarly, $f_3''$ represents the output of the second dense connection. $Bconvs$ is a multi-layer Conv-BN-ReLU layer, and $\odot$ represents element-wise multiplication. $Up_\uparrow^i$ is upsampling and represents an upsampling-concatenating strategy to integrate the multi-level features, where $i$ is the upsampling multiple. An overview of the GAM is provided in Fig. 4.

## D. Reverse-aware Module

The global map can only capture the relatively coarse position information for the roads, while ignoring the structural and textural details. To obtain more accurate feature information, we introduce the RAM, which is inspired by the reverse attention module [55]. This module discovers and identifies hidden areas by erasing specific parts of the objects. Specifically, this module can remove already estimated target areas to discover unnoticed complementary areas and details in the image. At the same time, this module is implemented through adaptive learning of four parallel features and can achieve implicit edge guidance without explicit boundary supervision shapes. However, unlike the original reverse attention module, which only removes the currently predicted foreground from the side output features and only focuses on the background features, the proposed RAM promotes the feature representation of both foreground and background simultaneously. An overview of the RAM is provided in Fig. 2.

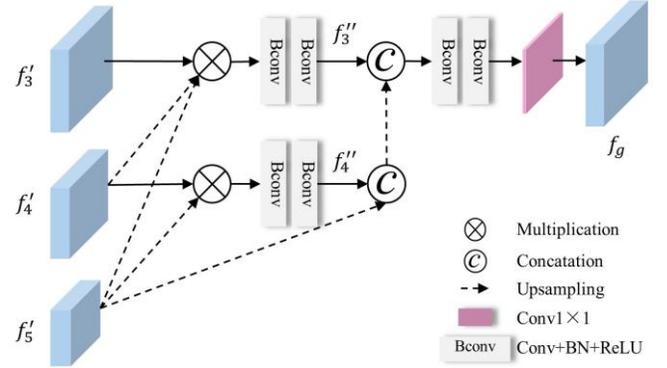

Fig. 4. The global aggregation module (GAM). The inputs are the three features of the positioning stage.

There are two separate paths included in the RAM. For the foreground path, the two inputs are directly connected along the channel dimension, and then sequential convolution, batch normalization, and ReLU activation layers are applied to obtain the foreground features $f_k^{fg}$. For the background path $f_k^{bg}$, we selectively focus on the background information. Specifically, complementary regions and details are sequentially mined by removing the predicted road areas from the side output features $f_{out}$, which can be expressed as:

$$f_k^{fg} = Conv\left(Bconvs\left(cat\left(Up_\uparrow^2(f_k), r_k\right)\right)\right) \quad (4)$$

$$f_k^{bg} = Conv\left(Bconvs\left((1-\sigma(r_k)) \odot f_k\right)\right) \quad (5)$$

$$f_{out} = f_k^{fg} + f_k^{bg} \quad (6)$$

where $Conv$ is a convolutional layer, $Bconvs$ is a multi-layer Conv-BN-ReLU layer, $cat$ is the concatenate operation, $Up_\uparrow^2$ represents a 2× upsampling operation, $\sigma$ is the sigmoid function, and $\odot$ denotes element-wise multiplication.

## E. Loss Functions

Road extraction can be considered as a binary semantic segmentation problem to distinguish roads from the background. A combination of binary cross-entropy (BCE) loss and dice coefficient loss is used for the loss function. This approach ensures accurate pixel classification and precise road boundary delineation, enhancing the overall segmentation performance. The combined loss function optimizes both aspects of the task. The overall loss is calculated as:

$$l_{BCE}(P,Y) = -\sum_{i=1}^{W}\sum_{j=1}^{H}\left[y_{ij} \cdot \log p_{ij} + (1-y_{ij}) \cdot \log(1-p_{ij})\right] \quad (7)$$

$$l_{Dice}(P,Y) = 1 - \frac{2 \cdot |P \cap Y|}{|P| + |Y|} \quad (8)$$

$$l = l_{BCE} + l_{Dice} \quad (9)$$

where $P$ and $Y$ are the predicted value and the real label, respectively; $W$ and $H$ are the width and height of the image, respectively; the output probability value at the pixel position is $p_{ij}$; and the corresponding label is $y_{ij}$.

In R2-Net, deep supervision is performed on the five side outputs (i.e., D0, D1, D2, D3, and D4), where D0 is the result obtained by concatenating these four lateral outputs along the channel dimension. Each image is upsampled to the same size as the ground-truth map G. Therefore, each loss function is a weighted sum of all the side outputs. The final loss is defined as:

$$L = \sum_{i=1}^{5} w_i l_i \quad (10)$$

where $w_i$ and $l_i$ are the overall loss and the weight of the $i$th side output, respectively. The appropriate value of $w_i$ is discussed in the discussion section.

This deeply supervised approach allows the model to receive feedback on prediction results at multiple levels, thereby improving the model's understanding and performance in the road segmentation task.

## IV. EXPERIMENTS

### A. Experimental Datasets

*1) WHU-RuR+ Dataset:* The WHU-*RuR+* road dataset is a large-scale remote sensing dataset specifically designed for rural road extraction, comprising 36,098 pairs of rural road samples with semantic labels from eight different countries. Among these, 18,103 pairs constitute the training set, while 17,995 pairs make up the testing set. Each sample in the dataset has a size of 1024 × 1024 pixels, with a resolution ranging from 0.3 to 0.8 meters per pixel. To evaluate the model's performance in different scenarios, the testing set is further divided into 9,084 pairs for the agricultural scene test set and 8,911 pairs for the complex rural scene test set.

*2) DeepGlobe Dataset:* The DeepGlobe road dataset [56] contains 6,226 samples of rural and urban roads from six different countries, including images and corresponding annotated road maps. The sample size is 1024 × 1024 pixels, and the resolution is 0.5 m/pixel. Following Batra, Singh, Pang, Basu, Jawahar and Paluri [57], we split the annotated samples into 4,696 samples for training and 1,530 samples for testing.

### B. Implementation Details

**Optimization details:** For all the deep learning-based methods, the training images were randomly cropped into 768 × 768 with data augmentation, which included horizontal flipping, vertical flipping, and diagonal flipping. We set the threshold to 0.5 for generating the final output. All the experiments were conducted using an NVIDIA TITAN RTX GPU (with 24 GB GPU memory).

For the experiments on the WHU-RuR+ dataset, the training set and the test set contained 18,103 images and 17,995 images, respectively. For the experiments on the DeepGlobe dataset, the training set and the test set contained 4696 images and 1530 images, respectively. All the models were trained with the Adam optimizer, and the batch size was 8 for 120 epochs. The learning rate started from 2e−4 and was divided by 5 at epochs {70, 90, 110}. In the prediction phase, there was no test time augmentation (TTA), to ensure a fair comparison with the published methods.

**Evaluation Metrics:** Road extraction from high-resolution satellite imagery is commonly approached as a semantic segmentation task, and evaluating the model performance often involves using the intersection over union (IoU) and F1-score. The IoU, as expressed in Equation (11), quantifies the overlap between the predicted and ground-truth road segments:

$$IOU = \frac{TP}{TP + FP + FN} \quad (11)$$

where $TP$ represents the number of pixels correctly extracted as road, $FP$ represents the number of pixels from other objects incorrectly extracted as road, and $FN$ represents the number of road pixels incorrectly extracted as other objects.

The F1-score, as provided in Equation (12), serves as a key metric for evaluating the performance of extraction networks. It represents the weighted average of the precision and the recall, as defined in Equation (13) and Equation (14), respectively.

$$F1 = 2 \times \frac{Precision \times Recall}{Precision + Recall} \quad (12)$$

$$Precision = \frac{TP}{TP + FP} \quad (13)$$

$$Recall = \frac{TP}{TP + FN} \quad (14)$$

**Comparison algorithms:** To verify the superiority of the proposed R2-Net framework, the comparison algorithms include state-of-the-art road extraction models, among which U-Net [58] performs well in many image segmentation tasks, making it a reliable benchmark for road extraction tasks. D-LinkNet [19] not only won the DeepGlobe Road Extraction

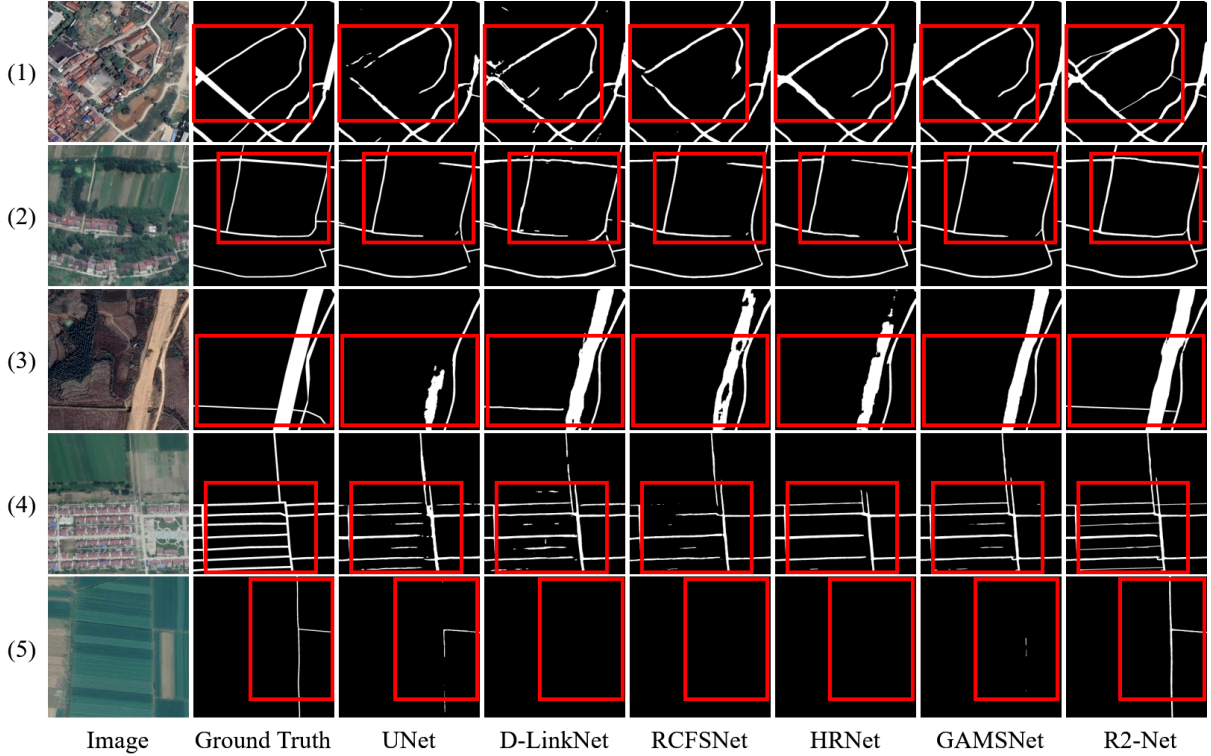

Fig. 5 Visual comparison of the different methods on the WHU-RuR+ Road Extraction dataset.

Challenge but also advanced the field of road extraction in terms of performance and innovation. HRNet [59] achieves high-precision image segmentation by maintaining high-resolution features and fusing multi-scale information. RCFSNet [21] is a network that combines road context information and full-stage feature fusion, while GAMSNet [20] is a globally aware road extraction network based on multi-scale residual learning to significantly improve the ability of road extraction.

*C. Comparison Experiments on the WHU-RuR+ Dataset*

This section illustrates the effectiveness of the proposed R2-Net framework through experimental analysis on the WHU-RuR+ dataset. Fig. 5 shows five representative visual results of WHU-RuR+, including rural scenes and farmland scenes. Among them, general rural scenes usually include rural towns, villages, forests, mountains, and other environments. The roads may be narrow and complex. The difficulty lies in the complex background of the road, which leads to serious occlusion and other phenomena. The background of the farmland scene is mainly dominated by large areas of farmland. The roads are usually straight and the width may be relatively narrow. The material is mostly dirt roads or simple paved roads. The difficulty of road extraction is that the road is often not clearly distinguished from the background, resulting in serious missed detection. The first row shows rural scenes, which usually have narrow roads and complex backgrounds. Most methods failed to fully extract the road, while R2-Net achieved more comprehensive recognition in this context, showing its robustness in dealing with complex backgrounds. The second row shows the case of occluded roads. Many methods resulted in fragmented extraction results due to occlusion, but R2-Net successfully recovered the features of occluded roads with its multi-scale approach and context-aware structure. The third row focuses on irregular road boundaries, especially on unpaved roads. Most methods failed to extract clear boundaries, showing their limitations under complex shapes, while R2-Net provided relatively clear road boundaries. The fourth and fifth rows involve scenes with thin and narrow roads, in rural areas and farmlands, respectively. The visualization results show that most of the comparison methods have a break when dealing with these narrow roads, mainly due to their lack of sensitivity in the feature extraction process. In contrast, R2-Net performs better in the extraction results of narrow roads. Table I lists the quantitative results on the WHU-RuR+ test set, covering the performance of rural scenes, farmland scenes, and the overall test set. R2Net has shown significant superiority over other methods on the WHU-RuR+ dataset. In the rural scene test, the recall rate of R2Net reached 74.04%, and the F1-score was 70.40%, both higher than the comparative methods, showing its advantages in capturing information about narrow rural roads in complex backgrounds. The Road IoU also reached 54.32%, indicating strong capability in accurately locating roads. In the farmland scene, the recall rate of R2Net is 73.31%. Although the F1-score drops slightly, it is still better than other methods, showing its effectiveness in the agricultural context. On the overall test set, R2Net once again performed well, with a recall rate of 73.79% and an F1-score of 69.40%, surpassing all comparison methods and demonstrating its robustness and effectiveness in a variety of scenarios. The road IoU is 53.14%, which is also the highest. Overall, R2Net can effectively meet the challenges of road extraction in rural and farmland environments, especially in complex backgrounds and narrow road extraction capabilities. It has strong application potential and can provide important data support for related fields.

TABLE I
THE QUANTITATIVE RESULTS ON THE WHU-RuR+ DATASET

| Test Set | Method | Recall (%) | F1-score (%) | Road IoU (%) |
|---|---|---|---|---|
| Test set1: General rural road scene | U-Net | 68.58 | 67.59 | 51.04 |
| | D-LinkNet | 68.94 | 69.32 | 53.05 |
| | HRNet | 68.49 | 68.93 | 52.59 |
| | GAMSNet | 69.33 | 69.26 | 52.98 |
| | RCFSNet | 64.17 | 68.03 | 51.55 |
| | R2-Net | **74.04** | **70.40** | **54.32** |
| Test set2: Farmland road scene | U-Net | 69.10 | 65.47 | 48.66 |
| | D-LinkNet | 71.80 | 67.64 | 51.10 |
| | HRNet | 67.98 | 65.83 | 49.07 |
| | GAMSNet | 67.52 | 66.33 | 49.62 |
| | RCFSNet | 67.15 | 66.52 | 49.84 |
| | R2-Net | **73.31** | **68.15** | **51.69** |
| Total test set | U-Net | 68.76 | 66.83 | 50.18 |
| | D-LinkNet | 68.49 | 68.93 | 52.59 |
| | HRNet | 66.35 | 68.10 | 51.63 |
| | GAMSNet | 69.60 | 68.76 | 52.40 |
| | RCFSNet | 65.21 | 67.48 | 50.92 |
| | R2-Net | **73.79** | **69.40** | **53.14** |

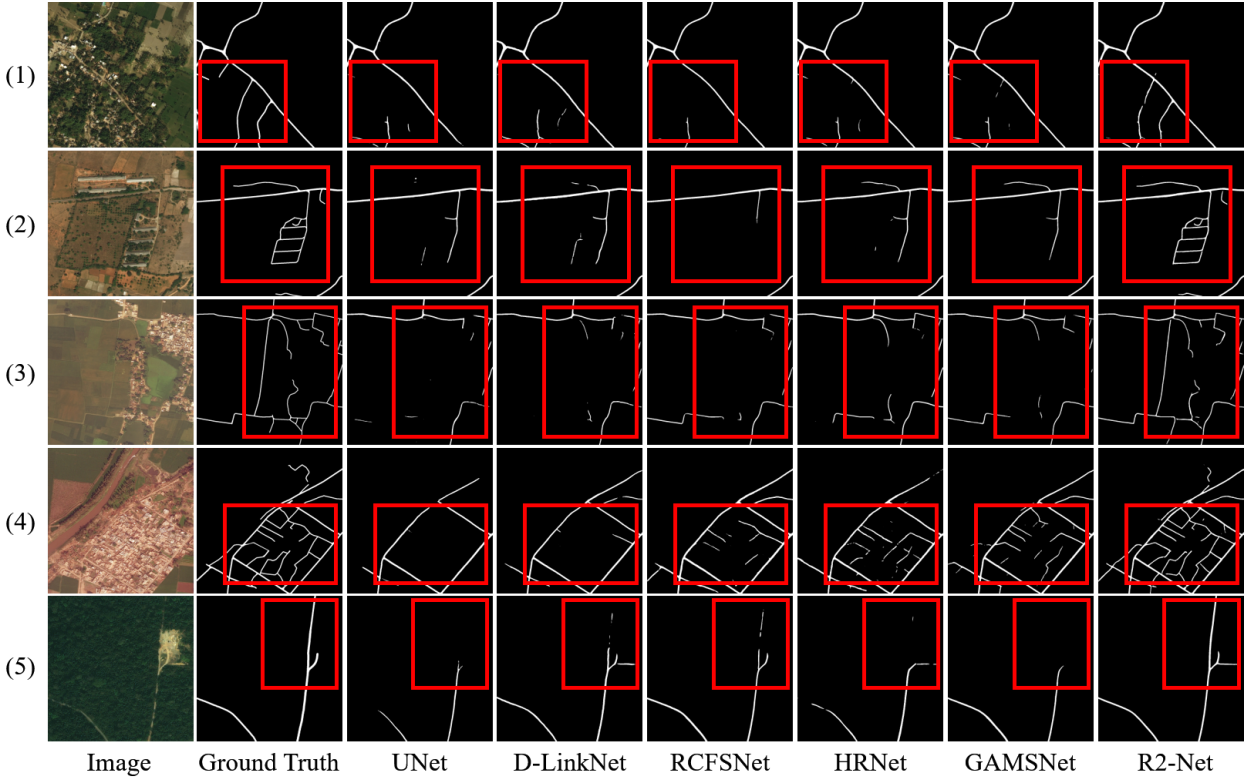

Fig. 6 Visual comparison of the different methods on the DeepGlobe Road Extraction dataset.

*D. Comparison Experiments on the DeepGlobe Dataset*

This section illustrates the effectiveness of the proposed R2-Net framework through experimental analysis on the DeepGlobe dataset. Fig. 6 shows five representative visual results of DeepGlobe. In the first row, the road is occluded by trees, making road extraction challenging. The proposed R2-Net method performs well and can effectively extract the occluded road information. In the scene shown in the second row, the similarity of spectral characteristics between the road and the background poses a huge challenge. Many compared methods can only extract the basic skeleton of the road network, resulting in a large number of omissions in narrow road sections. In contrast, when dealing with such spectrally highly similar scenes, the proposed R2-Net method outperforms other methods and can extract road information in more detail and reduce omissions. The third and fourth rows show a densely populated residential scene, where the texture of the narrow road is very similar to that of the surrounding buildings. The road is severely occluded by buildings. Compared with other methods, the proposed method can capture road features more finely and effectively overcome the challenge of road omissions in densely populated residential areas. The fifth row is a farmland scene, where the road is so narrow that most methods cannot accurately extract the complete road, while r2net effectively overcomes this challenge. Overall, the visualization results highlight the superior performance of the proposed R2-Net method in a variety of challenging scenes, demonstrating its robustness in extracting narrow roads in different environments. Table II lists the quantitative results on the DeepGlobe test set. R2-Net performs well in terms of recall, F1-

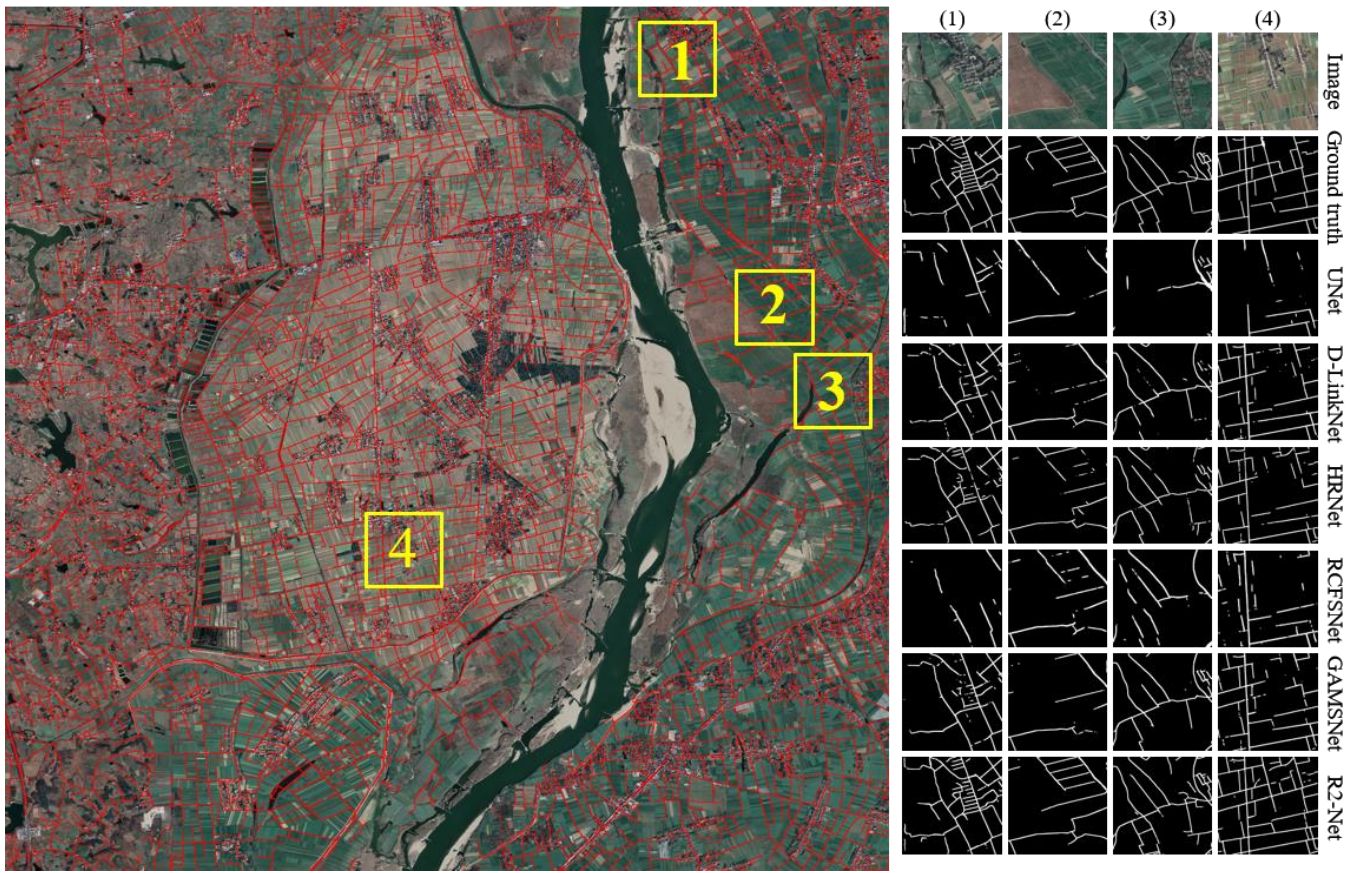

Fig. 7 The large-scale visualization results of R2-Net on Region A and the visualization results of the comparison methods in five random areas. Region A covers a large area of farmland, villages, etc., where there are many narrow rural roads.

score and road IoU, achieving 85.45%, 81.73% and 69.12%, respectively. In terms of IoU, there is no obvious gap between the proposed method R2-Net and GAMSNet, which may be because R2-Net focuses mainly on rural roads, while the DeepGlobe dataset contains a large number of urban roads, and GAMSNet is more suitable for this scenario. However, compared with GAMSNet, the proposed R2-Net method shows an improvement of 4.31% in recall, which further validates the effectiveness of the proposed method in comprehensively extracting the road sections in the image, and especially in identifying and capturing road details.

*E. Large Scale Image Inference*

Fig. 7 shows the visualization results of the proposed R2-Net method and the comparison methods on Regions A, which covers the cities of Zhongxiang and Jingmen, located in the Jianghan Plain in central China. The region covers a large area of farmland, villages, etc. The resolution of this area is 0.8 m/pixel, with a coverage of approximately 236 km². To visually compare the different methods, we randomly zoom in on five local areas to compare the details in the figure, where the size of each area is 1280 × 1280 pixels. Subfigure (a) is the optical image and subfigure (b) is the ground truth.

Region A covers a vast area of farmland and villages, with many scenes with similar road and background spectra. Rural roads are often unhardened roads. The spectral and shape characteristics of such roads and farmland background are not significant, and it is easy to misjudge roads as non-road areas. There are many missed extractions when the comparison methods deal with the narrow rural roads in the scenes with similar background spectra. Among the five comparison methods, U-Net obtains the worst results, often showing missed extraction of the narrow roads. D-LinkNet-50, RCFSNet, and GAMSNet perform better in identifying the obvious roads but still misclassify roads as non-road areas in the rural scenes where the spectral and shape features are not significant. For example, as shown in the small picture in Fig. 7, the comparison methods misclassify roads into non-road areas, such as bare land or vegetation, to varying degrees. However, R2-Net can accurately extract the narrow rural roads in these challenging

TABLE II
THE QUANTITATIVE RESULTS ON THE DEEPGLOBE DATASET

| Method | Metric | | |
|---|---|---|---|
| | Recall (%) | F1-score (%) | Road IoU (%) |
| UNet | 76.17 | 77.24 | 62.92 |
| D-LinkNet | 78.67 | 81.01 | 67.79 |
| HRNet | 79.71 | 81.07 | 68.17 |
| GAMSNet | 80.13 | 81.23 | 68.78 |
| RCFSNet | 79.12 | 80.95 | 67.99 |
| R2-Net | **84.44** | **81.69** | **69.05** |

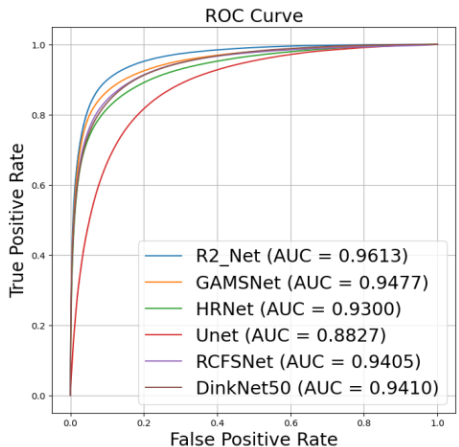

Fig. 8 The ROC curve for the large-scale image.

TABLE III
THE QUANTITATIVE RESULTS ON LARGE-SCALE IMAGE

| Method | Metric | |
|---|---|---|
| | F1-score (%) | Road IoU (%) |
| UNet | 38.67 | 55.77 |
| D-LinkNet | 41.22 | 58.38 |
| HRNet | 40.98 | 58.14 |
| GAMSNet | 41.86 | 59.02 |
| RCFSNet | 40.74 | 57.89 |
| R2-Net | **43.05** | **60.19** |

scenes and is less susceptible to interference from road and background spectral similarities.

To evaluate the quality of the large-scale rural road mapping, we show the receiver operating characteristic (ROC) curves for region A in Fig. 8. The six curves in each subfigure correspond to the six methods. Overall, the proposed R2-Net method significantly outperforms the other five comparison methods in the ROC curves in these three regions. Specifically, the ROC curve of the R2-Net model is the steepest, with the area under the ROC curve (AUC) reaching 0.9613, indicating that it achieves a high accuracy in road extraction. In addition to ROC curve comparisons, we also evaluated the quantitative results of the different methods on the large images using the road IoU and road F1-score. The evaluation results are presented in Table III, where the rows represent the tested methods and the columns show the results of the study area. R2-Net consistently performs exceptionally well in all the test areas, demonstrating higher IoU and F1-score values. This indicates that R2-Net has a significant performance advantage for large-scale rural road mapping applications.

*F. Ablation Study*

In this section, we study the impact of each module on the road extraction performance and conduct experimental analysis on the weight of the loss function. All the experiments were conducted on the WHU-RuR+ road dataset.

*1) Influence of the three modules:* In this section, we describe the ablation study conducted on the WHU-RuR+ rural road dataset, using ResNeSt-50 as the backbone and incrementally adding the individual modules. The three core modules were subject to ablation experiments. Among the different core modules, the ACAM is used to capture the long-distance spatial context information in various layers. The GAM facilitates cross-layer contextual connections while maintaining semantic consistency within a single layer. The RAM directs the network's attention toward the complex background, thereby amplifying its separability. As shown in Table IV, incorporating the ACAM results in a 0.73% improvement in road IoU and a 0.64% improvement in F1-score, compared to the baseline, demonstrating its effectiveness in extracting roads under challenging conditions. Similarly, with the GAM, the road IoU increases by 0.88% and the F1-score increases by 0.77%. Introducing the RAM further improves the road IoU by 1.34% and the F1-score by 1.16%. To further visualize the effect of the RAM, we visualized its feature maps, as shown in Fig. 9. The first row shows the input images, the second row shows the feature maps without the RAM, and the third row shows the feature maps with the RAM. It is evident that the inclusion of this module significantly enhances the distinction between foreground roads and the background. Each component in the ablation experiments impacted the model's performance, confirming their necessity throughout the model and reinforcing the comprehensiveness and effectiveness of the proposed model in the rural road extraction task.

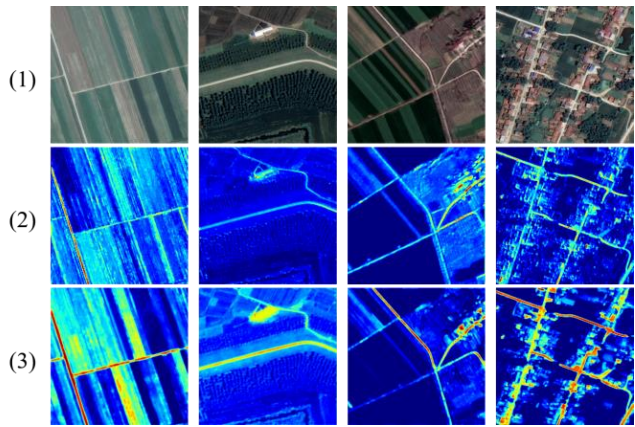

Fig. 9. Visualization of the feature maps: (1) input images, (2) without the RAM, and (3) with the RAM.

TABLE IV
ABLATION STUDY FOR THE MODULES OF R2NET

| Modules | | | | Metrics | |
|---|---|---|---|---|---|
| Baseline | ACAM | GAM | RAM | Road IoU (%) | F1 (%) |
| √ | | | | 50.50 | 67.11 |
| √ | √ | | | 51.23 | 67.75 |
| √ | √ | √ | | 52.11 | 68.52 |
| √ | √ | | √ | 52.57 | 68.91 |
| √ | √ | √ | √ | **53.14** | **69.40** |

TABLE V
ABLATION STUDY FOR THE LOSS FUNCTION WEIGHTS

| ID | Weight of side output | | | | | Metrics | |
|---|---|---|---|---|---|---|---|
| | D0 | D1 | D2 | D3 | D4 | Road IoU (%) | F1 (%) |
| 1 | 1 | 0 | 0 | 0 | 0 | 52.59 | 68.93 |
| 2 | 1 | 1 | 1 | 1 | 1 | 52.71 | 69.03 |
| 3 | 1.3 | 0.7 | 0.7 | 0.7 | 0.7 | 52.94 | 69.23 |
| 4 | 1.3 | 0.7 | 0.7 | 0.7 | 1 | 53.09 | 69.36 |
| 5 | 1.3 | 1 | 0.7 | 0.7 | 1 | **53.14** | **69.40** |

*2) Influence of weights of the side output layers*: R2-Net performs deep supervision through five side outputs. The specific structure is shown in Fig. 2. We adopted different weight settings for each side output in the experiments to evaluate the impact of deep supervision on road extraction performance. The quantitative comparison with the WHU-RuR+ rural road dataset is presented in Table V. The first column of Table V corresponds to the experiment ID for the different parameter settings.

In Experiment 1, we did not employ a deep supervision strategy for the side output layers. In Experiment 2, all five side output layers were assigned equal weights, while in Experiment 3, the highest weight was assigned to the last side output. Experiments 4 to 5 involved different weight adjustments for the side output layers. The results indicate that, compared to Experiment 1 without deep supervision, the other experiments showed improvements in all the evaluation metrics, suggesting a positive impact of the deep supervision on the model training. Upon further examination from Experiment 2 to Experiment 5, we observed that the performance was better when different weights were assigned to the side output layers, compared to the equal-weight setting. This can be attributed to the flexibility gained during training when allocating distinct weights to each side output layer. This allows the model to focus more dynamically on different levels of feature representation, leading to a more accurate understanding of the various aspects of roads and, consequently, an enhancement in overall performance. Through quantitative comparative experiments, we assessed the impact of different weight settings on the road extraction performance, ultimately selecting Experiment 5 as our optimal deep supervision strategy.

## V. CONCLUSION

In this article, we have proposed the reverse refinement network (R2-Net), which is a novel framework specifically designed for the extraction of narrow rural roads. Since rural road extraction faces many challenges, including narrow roads, complex backgrounds, spectral similarity, etc., existing methods often have difficulty capturing the details of these roads, resulting in incomplete or inaccurate extraction results. To address this issue, R2-Net cleverly integrates low-level detail information in the feature extraction stage, high-level semantic information in the localization stage, and global context information in the refinement stage. Specifically, by introducing the Axis Context Aware Module (ACAM), R2-Net can effectively capture context information at different scales, thereby improving the recognition ability of road features. In addition, the introduction of the reverse-aware module (RAM) further enhances the integration of foreground and background information, significantly improving the separability of road and non-road areas. In experiments, R2-Net performed well on multiple public data sets such as WHU-RuR+ and DeepGlobe, especially when dealing with narrow roads and complex backgrounds, successfully reducing missed mentions and false mentions. In addition, we verified it on a large scale and compared with other methods, R2-Net significantly improved the accuracy of road extraction, which shows that R2-Net has significant advantages for large-scale rural road mapping applications.

In the future, we will continue to advance research in this area by strengthening neural network architectures specifically designed for rural road extraction, improving road topology construction methods, and enhancing the generalization capability of road extraction models. Our goal is to further increase the accuracy and applicability of rural road mapping techniques, thereby contributing to more effective rural road research globally.